\documentclass[utf8]{frontiersSCNS} % for Science, Engineering and Humanities and Social Sciences articles
\usepackage{url,hyperref,lineno,microtype,subcaption}
\usepackage[onehalfspacing]{setspace}
\usepackage{todonotes}

\usepackage[a4paper,margin=2cm]{geometry}
\usepackage{graphicx}
\usepackage{tikz}
\usetikzlibrary{trees,positioning,shapes,shadows,arrows}

\usepackage{lipsum}  
\usepackage{siunitx} 
\usepackage{booktabs}

\usepackage[dvipsnames]{xcolor}
\colorlet{BLUE}{blue}
\colorlet{BLACK}{black}

\usepackage{multirow, makecell}
\usepackage{rotating}

\usepackage[nolist]{acronym} % acronyms by the \ac{label} command

\newacro{vic}[VIC]{Variable Impedance Control}
\newacro{pi2}[PI\textsuperscript{2}]{Policy Improvement with Path Integrals}
\newacro{dmp}[DMP]{Dynamic Movement Primitive}
\newacro{seds}[SEDS]{Stable Estimator of Dynamical Systems}
\newacro{ilc}[ILC]{Iterative  learning  control}
\newacro{gmm}[GMM]{Gaussian Mixture Model}
\newacro{gmr}[GMR]{Gaussian Mixture Regression}
\newacro{ppc}[PPC]{Passivity-Preservation Control}
\newacro{tpgmm}[TP-GMM]{Task-Parameterized \ac{gmm}}
\newacro{lfd}[LfD]{Learning from Demonstration}
\newacro{il}[IL]{Imitation Learning}
\newacro{wls}[WLS]{weighted least-squares}
\newacro{spd}[SPD]{Symmetric Positive Definite}
\newacro{emg}[EMG]{Electromyography}
\newacro{vil}[VIL]{Variable Impedance Learning}
\newacro{vilc}[VILC]{Variable Impedance Learning Control}
\newacro{ai}[AI]{Artificial Intelligent}
\newacro{sea}[SEA]{Series Elastic Actuation}
\newacro{rl}[RL]{Reinforcement Learning}
\newcommand{\bm}[1]{\boldsymbol{\mathbf{#1}}}

\newcommand{\Kpt}{{\bm{K}_t^\mathcal{P}}}
\newcommand{\Dvt}{{\bm{D}_t^\mathcal{V}}}
\newcommand{\Kot}{{\bm{K}_t^\mathcal{O}}}
\newcommand{\Dwt}{{\bm{D}_t^\mathcal{W}}}

\newcommand{\Kp}{{\bm{K}^\mathcal{P}}}
\newcommand{\Dv}{{\bm{D}^\mathcal{V}}}
\newcommand{\Ko}{{\bm{K}^\mathcal{O}}}
\newcommand{\Dw}{{\bm{D}^\mathcal{W}}}

\newcommand{\xhat}{{\bm{\hat{x}}}}
\newcommand{\fet}{{\bm{f}_t^e}}
\newcommand{\tauet}{{\bm{\tau}_t^e}}
\newcommand{\spd}{{\bm{\mathcal{S}}_+^m}}

\newcommand{\trsp}{{^{\top}}}

\providecommand{\change }[1]{{\color{black}#1}}
\newenvironment{change2}{%
	\color{black}
}{}

\providecommand{\Change }[1]{{\color{black}#1}}

\newcommand{\figref}[1]{\hyperref[#1]{Fig.~\ref*{#1}}}
\newcommand{\Figref}[1]{\hyperref[#1]{Figure~\ref*{#1}}}

\newcommand{\tabref}[1]{\hyperref[#1]{Table~\ref*{#1}}}
\newcommand{\secref}[1]{\hyperref[#1]{Section~\ref*{#1}}}
\newcommand{\algoref}[1]{\hyperref[#1]{Algorithm~\ref*{#1}}}

\newcommand{\eg} {\textit{e.g.,}~} %
\newcommand{\etc}{\textit{etc}} %

\newlength{\Oldarrayrulewidth}
\newcommand{\Cline}[2]{%
	\noalign{\global\setlength{\Oldarrayrulewidth}{\arrayrulewidth}}%
	\noalign{\global\setlength{\arrayrulewidth}{#1}}\cline{#2}%
	\noalign{\global\setlength{\arrayrulewidth}{\Oldarrayrulewidth}}
}

\usepackage{threeparttable, tablefootnote}

\def\keyFont{\fontsize{8}{11}\helveticabold }
\def\firstAuthorLast{Sample {et~al.}} %use et al only if is more than 1 author
\def\Authors{Fares J. Abu-Dakka\,$^{1,*}$ and Matteo Saveriano\,$^{2}$} % and Co-Author\,$^{1,2}$}
\def\Address{$^{1}$Intelligent Robotics Group, Department of Electrical Engineering and Automation (EEA), Aalto University, Espoo, Finland \\
	$^{2}$Intelligent and Interactive Systems, Department of Computer Science and Digital Science Center (DiSC), University of Innsbruck, Innsbruck, Austria}
\def\corrAuthor{Fares J. Abu-Dakka}

\def\corrEmail{fares.abu-dakka@aalto.fi}

\begin{document}
	\onecolumn
	\firstpage{1}
	
	\title[VILC - A review]{Variable impedance control and learning -- A review} 
	
	\author[\firstAuthorLast ]{\Authors} %This field will be automatically populated
	\address{} %This field will be automatically populated
	\correspondance{} %This field will be automatically populated
	
	\extraAuth{}% If there are more than 1 corresponding author, comment this line and uncomment the next one.
	%\extraAuth{corresponding Author2 \\ Laboratory X2, Institute X2, Department X2, Organization X2, Street X2, City X2 , State XX2 (only USA, Canada and Australia), Zip Code2, X2 Country X2, email2@uni2.edu}

	\maketitle

	\begin{abstract}
		
		Robots that physically interact with their surroundings, in order to accomplish some tasks or assist humans in their activities, require to exploit contact forces in a safe and proficient manner.
		Impedance control is considered as a prominent approach in robotics to avoid large impact forces while operating in unstructured environments. In such environments, the conditions under which the interaction occurs may significantly vary during the task execution. This demands robots to be endowed with on-line adaptation capabilities to cope with sudden and unexpected changes in the environment. In this context, variable impedance control arises as a powerful tool to modulate the robot's behavior in response to variations in its surroundings. In this survey, we present the state-of-the-art of approaches devoted to variable impedance control from control and learning perspectives (separately and jointly). Moreover, we propose a new taxonomy for mechanical impedance based on variability, learning, and control. The objective of this survey is to put together the concepts and efforts that have been done so far in this field, and to describe advantages and disadvantages of each approach. The survey concludes with open issues in the field and an envisioned framework that may potentially solve them. 
		
		%For full guidelines regarding your manuscript please refer to \href{http://www.frontiersin.org/about/AuthorGuidelines}{Author Guidelines}.
		%Leave the Abstract empty if your article does not require one, please see \href{http://www.frontiersin.org/about/AuthorGuidelines#SummaryTable}{Summary Table} for details according to article type. 

		\tiny
		\keyFont{ \section{Keywords:} impedance control, variable impedance control, variable impedance learning, variable impedance learning and control, variable stiffness} %All article types: you may provide up to 8 keywords; at least 5 are mandatory.
	\end{abstract}

\section{Introduction}
\label{sec:introduction}

Day by day realistic applications (e.g., disaster response, services and logistics applications, etc.) are bringing robots into unstructured environments (e.g., houses, hospitals, museums, etc.) where they are expected to perform complex manipulation tasks. This growth in robot applications and technologies is changing the classical view of robots as caged manipulators in industrial settings.
Indeed, robots are now required to directly interact with unstructured environments which are dynamic, uncertain, and possibly inhabited by humans. This demands to use advanced interaction methodologies based on impedance control.

Classical robotics, mostly characterized by high gain negative error feedback control, which is not suitable for tasks that involve interaction with the environment (possibly humans), because of possible high impact forces. The use of impedance control provides a feasible solution to overcome position uncertainties and subsequently avoid large impact forces, since robots are controlled to modulate their motion or compliance according to force perceptions. 
Note that compliance control \citep{salisbury1980active} is only a subset of impedance control to produce compliant motion \citep{Park2019} and its tradition definition is ``any robot motion during which the end-effector trajectory is modified, or even generated, based on online sensor
information'' \citep{de1987study}. 
Noteworthy, in cases where a robotic system is not providing access to low-level control (\eg commanding joints torque or motor current), then we model interaction between the robot and the environment using admittance control \citep{Villani2016} by translating the contact forces into velocity commands\footnote{Admittance control maps, in a physically consistent manner, sensed external forces into desired robot velocities. Therefore, it can be considered as the opposite, or the dual, of the impedance control. Unlike impedance control, admittance Control performs more accurate execution in non-contact tasks or even in contact with non-stiff (viscous) environments. \change{In practice, the choice between impedance and admittance control often depends on the available robot. It is known that, to realize an impedance controller, one has to directly control the joint torques (or the motor current). However, most of the available robotic platforms are controlled in position or velocity and the roboticists has no access to the force/torque level. This is especially true for industrial robots. In this cases, admittance control is an effective strategy to realize a desired impedance behavior on a position (velocity) controlled robots. The user has to equip the robot with additional sensors, like a force/torque sensor at the end-effector, and convert the measured forces into desired position (velocity) commands.}
	
	\cite{ott2010unified} proposed a hybrid system that incorporates both controllers in order to achieve: (\emph{i}) accuracy in free motion and (\emph{ii}) robust interaction while in-contact and free of impacts. \change{Recently, \cite{Bitz2020Variable} investigated the trade-off between agility and stability by proposing a variable damping controller based on user intent. }
	Admittance control is out of the scope of this review and will be partially mentioned. The interested reader is referred to \citep{Keemink2018admittance} for a review on admittance control-based techniques applied to human-robot interaction.}.

\begin{figure}[t]
	\centering
	\includegraphics[width=0.9\columnwidth]{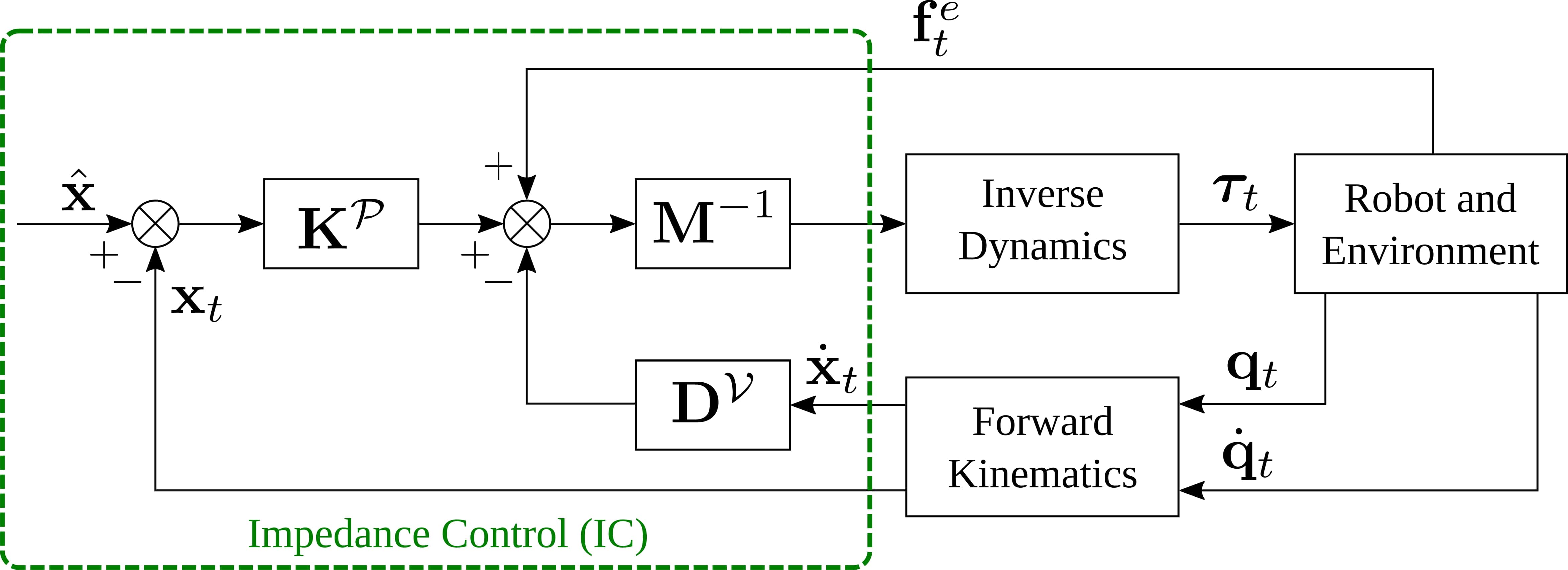}
	\caption{\change{Block scheme of impedance control, obtained assuming that $\dot{\hat{\mathbf{x}}} = \ddot{\hat{\mathbf{x}}}  = \mathbf{0}$.}}
	\label{fig:impedance_control}
\end{figure}

Impedance controller resembles a virtual spring–damper system between the environment and robot end-effector \citep{Hogan1985}, which allows robots to interact with the environment or humans more safely and in an energy-efficient way. 
In learning applications, a standard impedance interaction model is defined as
\begin{change2}
	\begin{align}
	\bm{M}\bm{\ddot{x}}_t &= \Kp(\xhat-\bm{x}_t)-\Dv\bm{\dot{x}}_t+\fet, &(&translational)
	\label{eq:msd_trans} \\
	\bm{I}\bm{\dot{\omega}}_t &= \Ko(\text{log}(\bm{\hat{R}}\bm{R}_t\trsp))-\Dw\bm{\omega}_t+\tauet, &(&rotational)
	\label{eq:msd_ori}
	\end{align}
	
	where $t=1,2,\cdots,T$ is the time-step, $\xhat$ represents the goal position (or desired trajectory\footnote{The reader with a control background will recall that, in impedance control, the interaction model is defined in terms of the error $\bm{e}_t$ (e.g., for the position, $\bm{e}_t = \xhat-\bm{x}_t$) and its derivatives $\bm{\dot{e}}_t$ and $\bm{\ddot{e}}_t$. The impedance model in \eqref{eq:msd_trans}--\eqref{eq:msd_ori} is obtained by assuming $\bm{\dot{\hat x}}_t = \bm{\ddot{\hat x}}_t = \bm{0}$. i.e. we aim at reaching the fixed goal defined by $\xhat$ and $\bm{\hat{R}}$.}), and $\bm{x}_t$ is the actual trajectory of the end-effector. $\mathbf{M}, \Kp, \text{and } \Dv$ are the mass, stiffness and damping matrices, respectively, for translational motion, while $\mathbf{I}, \Ko, \text{and } \Dw$ are the moment of inertia, stiffness and damping matrices, respectively, for rotational motion. $\bm{\omega}_t$ is the angular velocity of the end-effector.
	$\bm{\hat{R}},\bm{R}_t\in\bm{SO}(3)$ are rotation matrices and correspond to desired rotation goal and actual orientation profile of the end-effector, respectively. The rotation of $\bm{R}_t$ into $\bm{\hat{R}}$  is defined as $\text{log}(\bm{\hat{R}}\bm{R}_t\trsp)$. $\fet$ and $\tauet$ represent the external force and torque applied to the robot end-effector. \change{\Figref{fig:impedance_control} shows the block scheme of the impedance control for the translational part.}
\end{change2}

Impedance control can be used in Cartesian space to control the end-effector interaction with the environment \citep{siciliano2001inverse,albu2002cartesian,lippiello2007position,caccavale2008six}, \change{like in haptic exploration \citep{eiband2019learning}}, as well as in joint space \citep{tsetserukou2009isora,li2011impedance,li2011model,li2012learning} to enhance safety.
\citeauthor{albu2003cartesian} studied Cartesian impedance control with null-space stiffness based on singular perturbation \citep{albu2003cartesian} and passive approach  \citep{albu2007unified}. Few years later, researches tackled null-space impedance control in multi-priority controllers \citep{sadeghian2011multi,hoffman2018multi} and to ensure the convergance of task-space error.  
\citep{Ott2008cartesian} described Cartesian impedance control and its pros and cons for torque controlled redundant robots.

Impedance control is not only of importance when robots interact with a stiff environment. As early mentioned, new robot applications are bringing robots to share human spaces that make the contact between them inevitable. In such situations, it is important to ensure the human safety \citep{goodrich2008human, Haddadin2013towards}. Impedance control plays an important role in human-robot interaction. These robots are not supposed to just be in human spaces to do some specific tasks, but also to assist human in many other tasks like lifting heavy objects (\eg table, box) \citep{ikeura1995, ikeura2002optimal}, objects handover \citep{Bohren2011towards, Medina2016human}, \etc, in human-robot collaboration framework \citep{bauer2008human}.

However, in many tasks robots need to vary their impedance along the execution of the task. As an illustrative example, robots in unstructured environments (homes, industrial floors or other similar scenarios) may require to turn valves or open doors, etc. Such tasks demand the application of different control forces according to different mass, friction forces, etc. In that sense, sensed forces convey relevant information regarding the control forces needed to perform such manipulation tasks, which can be governed through stiffness variations \citep{abu2018force}. Another example, from human-robot cooperative scenario, a robot needs to adapt its stiffness based on its interaction with a human in a cooperative assembly task \citep{Rozo2013}.

From now on the main focus of this survey will be on \emph{\ac{vic}} from both control and learning perspectives. To the best of our knowledge, this is the first survey that focuses on control and learning approaches for variable impedance. A thorough search of the relevant literature yielded to the list in \tabref{tab:sota}. %\Figref{fig:taxonomy} shows the proposed taxonomy of \ac{vic}.

\begin{table}[t]
	\centering
	\begin{tabular}{|m{0.18\textwidth}|m{0.14\textwidth}|m{0.6\textwidth}|}
		%\noalign{\hrule height 1.5pt}
		\Cline{3pt}{2-3}
		\multicolumn{1}{c|}{} & \multicolumn{1}{c|}{\textbf{Topic}} & \multicolumn{1}{c|}{\textbf{Description}} \\
		\noalign{\hrule height 1.5pt}
		\cite{Vanderborght2013variable} and \cite{wolf2016variable} & Variable impedance actuators & Realize \ac{vic} in hardware with dedicated elastic elements\tablefootnote{%
			\change{\ac{sea} are implemented by introducing intentional elasticity between the motor actuator and the load for robust force control, which subsequently improves saftey during the interaction of the robot with the environment \citep{Pratt1995Series}. \\
				\ac{sea} are out of the scope of this paper, however, interested readers can refer to \citep{Calanca2017Impedance} which summarizes the common controller architectures for \ac{sea}. \ac{sea} framework has been used in many applications, mainly in human--robot interaction scenarios. For instance, \cite{yu2013control, yu2015human} designed compliant actuators for gait rehabilitation robot and validated its controller in order to provide safety and stability during the interaction. \cite{Li2017Multi} proposed a multi-modal controller for exoskeleton rehabilitation robots, driven by \ac{sea}, that guarantee the stability of the system. For more recent approaches please refer 
				to \citep{Haninger2020Safe, Kim2020Position}. \\
				On the other hand, for recent achievments in \ac{vic} in soft robots, please refer to \citep{Ataka2020model,Sozer2020Pressure,Zhong2020novel,Li2020variable,Gandarias2020open}.}}. They reviewed all possibilities to create variable stiffness actuators and all main factors that influence the most common approaches. \\ \cline{1-3}
		\cite{calanca2015review} & Compliance control & Reviewed impedance and admittance controllers for both stiff and soft joint robots. \\ \cline{1-3}
		\cite{Keemink2018admittance} & Admittance control & Reviewed admittance controllers with a specific focus on human--robot interaction. \\ \cline{1-3}
		\cite{song2019tutorial}  & All above & %This review covered wider range of topics than the previous three. 
		This review compared hardware- and software-based approaches, and main technical developments about impedance control including hybrid impedance, force-tracking, and adaptive methods. However, learning algorithms and \ac{vic} methods are mentioned in two small subsections. \\ %\cline{1-3}
		\noalign{\hrule height 1.5pt}
		Our review  & \ac{vic}, \acs{vil}, \& \acs{vilc}   & This review departs from impedance control approaches to focus on learning and learning control approaches used to implement variable impedance behaviors. We analyze the advantages and disadvantages of traditional approaches based on control and recent frameworks that integrate learning techniques. Therefore, our review has a potential impact on both the control and the learning communities. \\ 
		\noalign{\hrule height 1.5pt}
	\end{tabular}
	\caption{Comparison between our review and the current reviews in the literature.}
	\label{tab:sota}
\end{table}

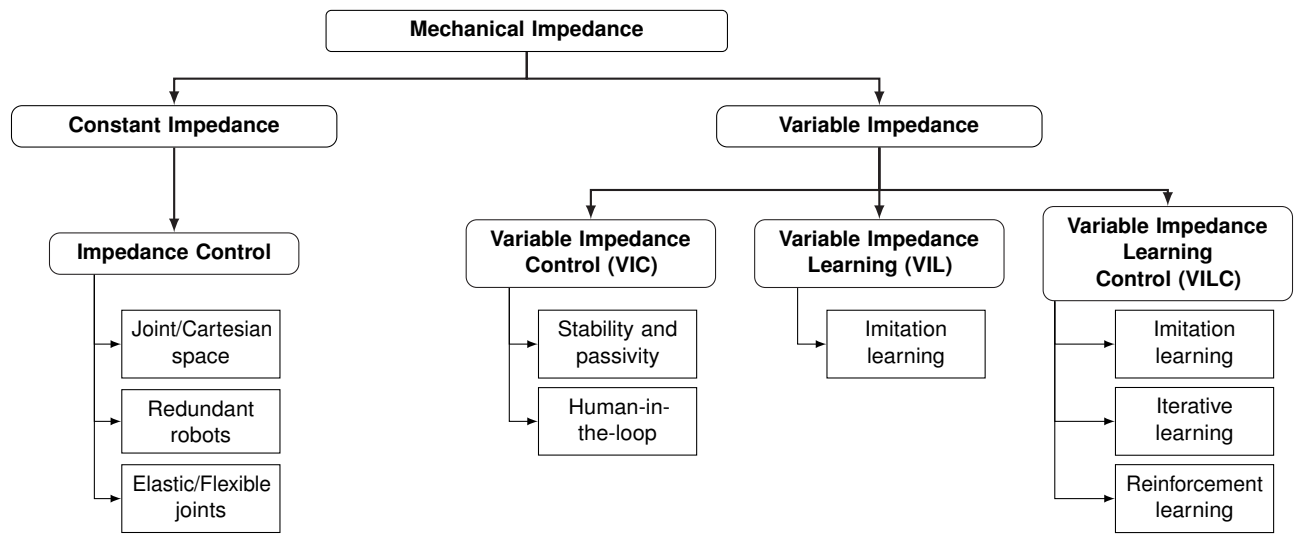
\begin{figure}[!th]
	\centering
	\tikzset{
		%basic/.style  = {draw, text width=5cm, drop shadow, font=\sffamily\footnotesize, rectangle}, % With shadow
		basic/.style  = {draw, text width=5cm, font=\sffamily\footnotesize, rectangle},
		root/.style   = {basic, rounded corners=2pt, thin, align=center, fill=white},
		level-2/.style = {basic, rounded corners=4pt, thin,align=center, fill=white, text width=4cm},
		level-3/.style = {basic, rounded corners=4pt, thin,align=center, fill=white, text width=3cm},
		level-4/.style = {basic, thin, align=center, fill=white, text width=1.8cm}
	}
	
	\begin{tikzpicture}[
	level 1/.style={sibling distance=22em, level distance=3em},
	level 2/.style={sibling distance=9em, level distance=4em},
	level 3/.style={sibling distance=4em, level distance=3em},
	%   {edge from parent fork down},
	edge from parent/.style={->,solid,black,thick,sloped,draw}, 
	edge from parent path={(\tikzparentnode.south) -- (\tikzchildnode.north)},
	>=latex, node distance=1.2cm, edge from parent fork down]
	
	% root of the the initial tree, level 1
	\node[root] {\textbf{Mechanical Impedance}}
	% The first level, as children of the initial tree
	child {node[level-2] (c1) {\textbf{Constant Impedance}}
		child {node[level-3] (c11) {\textbf{Impedance Control}}}}
	child {node[level-2] (c2) {\textbf{Variable Impedance}}
		child {node[level-3] (c21) {\textbf{Variable Impedance Control (VIC)}}}
		child {node[level-3] (c22) {\textbf{Variable Impedance Learning (VIL)}}}
		child {node[level-3] (c23) {\textbf{Variable Impedance Learning Control (VILC)}}}
	};
	%child {node[level-2] (c3) {\textbf{Category 3}}}
	%child {node[level-2] (c4) {\textbf{Category 4}}};
	
	% The second level, relatively positioned nodes
	\begin{scope}[every node/.style={level-4}]
	\node [below of = c11, xshift=10pt] (c111) {Joint/Cartesian space};
	\node [below of = c111, yshift=5pt] (c112) {Redundant robots};
	\node [below of = c112, yshift=5pt] (c113) {Elastic/Flexible joints};
	
	\node [below of = c21, xshift=10pt] (c211) {Stability and passivity};
	\node [below of = c211, yshift=5pt] (c212) {Human-in- the-loop};
	%\node [below of = c212, yshift=5pt] (c213) {Optimal control};
	
	\node [below of = c22, xshift=10pt] (c221) {Imitation learning};
	%\node [below of = c221, yshift=5pt] (c222) {Manifold learning};
	
	\node [below of = c23, xshift=10pt] (c231) {Imitation learning};
	\node [below of = c231, yshift=5pt] (c232) {Iterative learning};
	\node [below of = c232, yshift=5pt] (c233) {Reinforcement learning};
	\end{scope}
	
	% lines from each level 1 node to every one of its "children"
	\foreach \value in {1,...,3}
	\draw[->] (c11.195) |- (c11\value.west);
	
	\foreach \value in {1,2}
	\draw[->] (c21.203) |- (c21\value.west);
	
	\foreach \value in {1}
	\draw[->] (c22.203) |- (c22\value.west);
	
	\foreach \value in {1,2,3}
	\draw[->] (c23.203) |- (c23\value.west);
	\end{tikzpicture}
	\caption{A taxonomy of existing approaches for (variable) impedance learning and control.}
	\label{fig:taxonomy}
\end{figure}

\Figref{fig:taxonomy} shows the proposed taxonomy that categorize existing approaches in the field. Starting from the root we find the physical concept of \textit{mechanical impedance}. \Change{Mechanical impedance inspired preliminary work on impedance control~\citep{Hogan1985} where the key idea is to control the impedance behavior of a robotic manipulator to ensure physical compatibility with the environment. In impedance control, we identify two macro groups of approaches, namely those based on constant impedance gains and those based on variable impedance gains}. Standard impedance control is a way to actively impose a predefined impedance behavior on a mechanical system. It can be realized both with constant and variable impedance. Standard impedance control\footnote{Constant impedance is out of the scope of this paper, however, interested readers are advised to consult \citep{calanca2015review, song2019tutorial}} has been applied to control robots with rigid or elastic joints, both in joint and Cartesian spaces. The stability of a \ac{vic} scheme depends on how the impedance gains vary. Therefore, several approaches have been developed to investigate the stability of the controller eventually with a human-in-the-loop. The possibility of varying the impedance has been also investigated from the learning community. Approaches for \ac{vil} treat the problem of finding variable impedance gains as a supervised learning problem and exploit human demonstrations as training data (imitation learning). Typically, \ac{vil} approaches rely on existing controller to reproduce the learned impedance behavior. On the contrary, \ac{vilc} approaches attempt to directly learn a variable impedance control law. This is typically achieved via imitation, iterative, or reinforcement learning.     

\section{Variable Impedance Control \texorpdfstring{\change{(VIC)}}{}}
\label{sec:vic}

\begin{figure}[t]
	\centering
	\includegraphics[width=0.9\columnwidth]{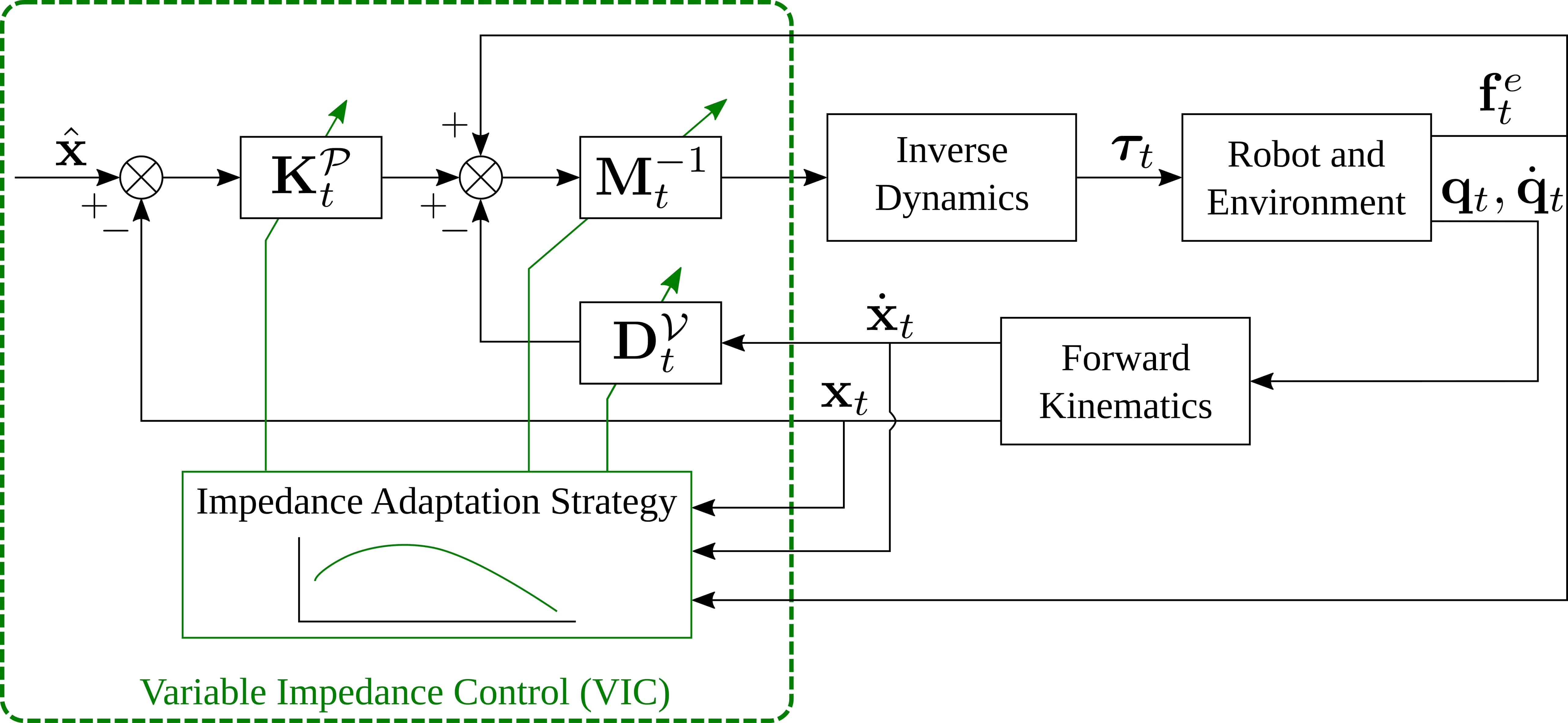}
	\caption{\change{Block scheme of \ac{vic}, obtained assuming that $\dot{\hat{\mathbf{x}}} = \ddot{\hat{\mathbf{x}}}  = \mathbf{0}$.}}
	\label{fig:variable_impedance_control}
\end{figure}

The increasing demand for robotic systems to assist human in industry, homes, hospitals, museums, \etc, has encouraged roboticists to investigate advanced interaction methods based on impedance control. In tasks that require a robot to interact physically with the environment (possibly human), impedance control provides a dynamic relationship between position and force in order to overcome any position uncertainties and subsequently avoid large impact forces. In the past decades, scholars have been investigated impedance control for a wide range of robot applications, \eg industry \citep{jung2004force}, \change{agricolture \citep{balatti2019towards}}, human-robot interaction \citep{magrini2015control}, and rehabilitation \citep{jamwal2016impedance}. However, since 1995 when \citeauthor{ikeura1995} proposed for the first time the concept of variable impedance control as a method for cooperative systems, researchers started massively to investigate \ac{vic}, in many robot applications, due to efficiency, flexibility and safety that can add to the systems controllers. 

\begin{change2}In order to write the standard formula for \ac{vic}, we need to slightly modify equations \eqref{eq:msd_trans} and \eqref{eq:msd_ori} into 
	
	\begin{align}
	\bm{M}_t\bm{\ddot{x}}_t &= \Kpt(\xhat-\bm{x}_t)-\Dvt\bm{\dot{x}}_t+\fet, &(&translational)
	\label{eq:msd_trans_vic} \\
	\bm{I}_t\bm{\dot{\omega}}_t &= \Kot(\text{log}(\bm{\hat{R}}\bm{R}_t\trsp))-\Dwt\bm{\omega}_t+\tauet, &(&rotational)
	\label{eq:msd_ori_vic}
	\end{align}
	where $\Kpt$, $\Dvt$, $\Kot$, and $\Dwt$ are the same quantities defined in \secref{sec:introduction}. The only difference is the subscript $t$ used to indicate that quantities are varying over time.
\end{change2}
\change{A block scheme that implements \ac{vic} is shown in \Figref{fig:variable_impedance_control}.}

\ac{vic} plays an important role in human-robot cooperation. One of the earliest works was introduced by
\cite{ikeura1995} to show the advantages of variable damping control schemes for a master–slave system to perform lifting tasks, which was then extended by introducing variable stiffness \citep{rahman1999}.
In \citeauthor{ikeura1995}'s system, the damping variation was estimated a priori (through experimental data), either using least-squares \citep{ikeura1995} or later by optimizing a suitable cost function \citep{ikeura2002optimal}. Later, \ac{vic} was used to provide a coordination mechanism between robot and human \citep{al1997arm}. \cite{tsumugiwa2002} introduced a variable impedance control based on the human arm stiffness estimation. They varied the virtual damping coefficient of the robot as a function of the estimated stiffness of a human arm, and differential changes in position and force. They used recursive least-squares method to estimate for the stiffness coefficient. \change{They applied signal processing with a digital filtering in order to overcome the influence of the measurement noise and subsequently improve the accuracy.}
In redundant manipulators, where robots are endowed with the ability to behave naturally as close as possible the desired impedance profile, \ac{vic} was used in a cooperative paining task \citep{Ficuciello2015variable}. Recently, a minimally model-based trajectory tracking \ac{vic} controller was proposed \citep{Spyrakos2020minimally}.

When two persons are collaborating or cooperating\footnote{In literature, authors are using collaboration and cooperation as synonymous. However, by consulting \citeauthor{merriamWebster}, \textit{Cooperation} means the actions of someone who is being helpful by doing what is wanted or asked, while \textit{Collaboration} means to work jointly with others or together especially in an intellectual endeavor. Based on this definition we will distinguish the use of the two words based on the intentions of the tasks. Normally robots are cooperating with human to help him to achieve something but not collaborating with him for a common goal.} to perform some task (\eg lift and transport a table), both can sense the intentions of the other partner by sensing the forces transmitted through the object and they act accordingly. However, when a robot is cooperating with a human, it is not obvious for the robot to understand human intention. \ac{vic} along with the estimation of human intentions have stimulated researchers efforts in the past couple of decades. \cite{duchaine2007general} estimated human intention by exploiting the time-derivative of the applied forces, in order to adjust damping parameter in the robot's controller. \change{Recently, \cite{muratore2019self} proposed a multimodal interaction framework that exploits force, motion, and verbal cues for human--robot cooperation. \ac{vic} is used to control the physical interaction and the joint stiffness is updated online using the simple adaptation rule 
	\begin{align}
	%k_{j,t} = k_{j,0} + \alpha \bm{e}_{j,t}^2, \quad j=1,\ldots,J\\
	\bm{k}_t = \bm{k}_0 + \alpha \, \bm{e}_t^2,
	\label{eq:stiffness_muratore}
	\end{align}
	where $\bm{k}_t = \text{diag}(\Kpt) = [k_{1,t},\ldots ,k_{j,t}, \ldots,k_{J,t}]\trsp$ is the joint stiffness and $J$ is the number of joints.
	%where $\bm{k}_t$ is the joint stiffness, 
	$\bm{k}_0= [k_{1,0},\ldots ,k_{j,0}, \ldots,k_{J,0}]\trsp$ is a small stiffness used to prevent unsafe interaction, $\bm{e}_t$ is the joint trajectory tracking error, and $\alpha$ is a positive gain. In practice, the update rule~\eqref{eq:stiffness_muratore} increases the stiffness of the joints with high error letting the robot to track more accurately the desired trajectory. }

In rehabilitation, \cite{blaya2004adaptive} implemented a \ac{vic} for an ankle-foot orthosis. However, they did not perform any stability analysis for their system. A \ac{vic} controller to provide a continuum of equilibria
along the gait cycle has been implemented \citep{mohammadi2019variable}. Stability analysis of their controller has been provided based on Lyapunov matrix
inequality. \change{Finally, \cite{arnold2019variable} proposed to control the ankle joint of a wearable exoskeleton robot to improve the trade-off between performance and stability.}

In grasping, \cite{ajoudani2016reflex} proposed a \ac{vic} controller to vary the stiffness based on the friction coefficient, estimated via exploratory action prior grasping, in order to avoid slippage of the grasped object. \change{In manipulation, \cite{johannsmeier2019framework} propose a framework that combines skills definition, stiffness adaptation, and adaptive force control. Similarly to \cite{muratore2019self}, the stiffness is updated considering the trajectory tracking error. The framework is evaluated on a series of peg-in-hole tasks with low tolerance ($< 0.1\,$mm) showing promising results.}

%In this section, we will discuss \ac{vic} from control perspective in different robotics applications, such as: (\emph{i}) human-robot cooperation, (\emph{ii}) rehabilitation.

\subsection{\texorpdfstring{\ac{vic}}{} stability \& passivity}
\label{sec:vic:stability}

Stability issues of impedance control has been studied from the beginning by \cite{Hogan1985} and later by \citep{colgate1988robust} where the passivity concept had been introduced. However, stability in \ac{vic} is not a trivial problem and has been recently considered in literature. One of the earliest stability analysis of \ac{vic} was for a force tracking impedance controller \citep{lee2008force}. In their controller, the target stiffness was adapted according to the previous force tracking error resulting in a second order linear time varying system.
\cite{ganesh2012versatile} implemented a \change{versatile bio-mimetic controller capable of automatic adjustment of the}  stiffness over a fixed reference trajectory while maintaining stability. 

Analyze the stability of an interaction with Lyapunov-based tools becomes hard when the dynamics of the environment are unknown. This is clearly the case of a robot physically interacting with a human operator. In this respect, \textit{passivity} arises as an intuitive way to investigate stability\footnote{Describe the passivity framework is out of the scope of this review. The interested user is referred to \citep{Van2000l2} for a thorough discussion on passivity theory.} Loosely speaking, the passivity theory introduces a mathematical framework to describe and verify the property of a dynamical system of not producing more energy than it receives.

A passivity-based approach is presented to ensure stability of a time-varying impedance controller \citep{Ferraguti2013tank}. They ensure the passivity by ensuring that the dissipated energy added to a virtual energy tank is greater than the energy pumped into the system. Their approach depends on the initial and threshold energy levels and on the robot state.
Later, \citeauthor{ferraguti2015energy} extended their approach to time-varying admittance controller in order to adapt the human movements where the passivity analysis took place using port-Hamiltonian representation \citep{ferraguti2015energy}. 
In contrast, \cite{Kronander2016stability} proposed state independent stability conditions for \ac{vic} scheme for varying stiffness and damping. They used a modified Lyapunov function for the derivation of the stability constraints for both damping and stiffness profiles. This idea of constraining variable impedance matrices to guarantee the stability on variable impedance dynamics before the execution has been expanded later by \cite{sun2019stability}. \citeauthor{sun2019stability} proposed new constraints to on variable impedance matrices that guarantee the exponential stability of the desired impedance dynamics while ensuring the boundedness of the robot's position,
velocity, and acceleration in the desired impedance dynamics.

Recently, \cite{Spyrakos2019Passivity} proposed a \ac{ppc} that enables the implementation of stable  \ac{vic}. They also provided joint and Cartesian space versions of the \ac{ppc} controller, to permit intuitive definition of interaction tasks. 

\subsection{\texorpdfstring{\ac{vic}}{} with human-in-the-loop}
\label{sec:vic:hloop}

Previous works have been devoted to understand how
impedance is modulated when humans interact with the environment \cite{Burdet2001} or to transfer human’s impedance-based skills to
robots \cite{Ajoudani2016}.  
The presence of the human, or human-in-the-loop, introduces a certain level of uncertainty in the system and poses several technical problems for the underlying controller that should guarantee stability of the interaction while effectively supporting the human during the task.
In this section we are covering potential research on \ac{vic} approaches from control perspective while having the human in the control loop.  However, robot learning capabilities to automatically vary impedance controller parameters to satisfactorily adapt in face of unseen situations while having human-in-the-loop will be covered in \secref{sec:vilc:hitl}.

In 2012, \citeauthor{ajoudani2012tele} introduced the concept of tele-impedance through a technique capable of transferring human skills in impedance (stiffness) regulation to robots (slave) interacting with uncertain environment. Human impedance where estimated in real-time using \ac{emg} to measure signals of eight muscles of human's arm (master). They applied this method to peg-in-hole application based on visual information and without any haptic feedback. \change{Few years later, the authors updated their result in \cite{laghi2020unifying} by overcoming the loss of transparency by integrating two-channel bilateral architecture with the tele-impedance paradigm.}
%, allowing the slave to follow not only the user's position and forces, but also its stiffness profiles.}

In control interfaces that include human-in-the-loop, \ac{emg} signals have been successfully used to estimate human impedance and subsequently use it as an input “intention estimate” for controlling robots in different tasks, \eg cooperative manipulation task \cite{peternel2016towards, peternel2017human, delpreto2019sharing}. \cite{peternel2016towards,peternel2017human} proposed a multi-modal interface, using \ac{emg} and force manipulability measurements of the human arm, to extract human's intention (stiffness behaviour) through muscles activities during cooperative tasks. Subsequently, a hybrid force/impedance controller uses the stiffness behaviour to perform the task cooperatively with the human.

\cite{Rahimi2018neural} propose a framework for human--robot collaboration composed of two nested control loops. The outer loop defines a target variable impedance behavior using a feed-forward neural network to adapt the desired impedance in order to minimize the human effort. The inner loop generates an adaptive torque command for the robot such that the unknown robot dynamics follows the target impedance behavior generated in the outer loop. An additive control term, approximated with a neural network whose weights are updated during the execution, is used to cope with the unknown dynamics deriving from the interaction. 

\section{Variable Impedance Learning \texorpdfstring{\change{(VIL)}}{}}
\label{sec:vil}

\begin{figure}[t]
	\centering
	\includegraphics[width=0.9\columnwidth]{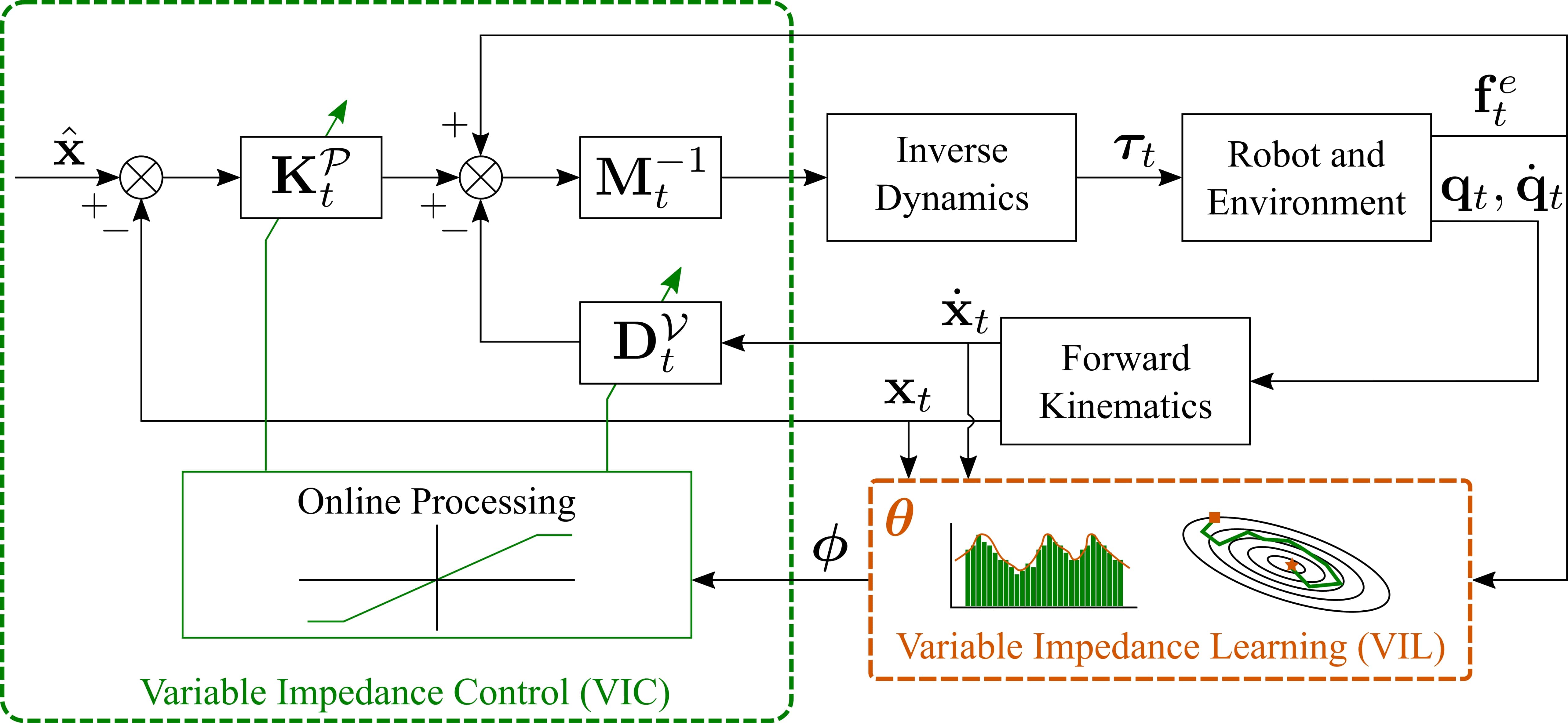}
	\caption{\change{Block scheme of  \ac{vil}, obtained assuming that $\dot{\hat{\mathbf{x}}} = \ddot{\hat{\mathbf{x}}}  = \mathbf{0}$.}}
	\label{fig:variable_impedance_learning}
\end{figure}

Traditionally, robot learning has been concerned about trajectory following tasks \citep{Ouyang2006}. However, the new generation of torque-controlled robots has made it possible to extend learning capabilities to tasks that require variable impedance skills \citep{abu2018force, abudakka2020}. Recently, robot learning algorithms have gained great interest for learning, reproducing, and adapting variable impedance parameters and treating them as skill to be learned. The focus in this section is on learning algorithms used to encode variable impedance gains for learning, reproduction, and adaptation, regardless of the effect of these gains on the robot behavior. The latter will be discussed in \secref{sec:vilc}. In our taxonomy (see \Figref{fig:taxonomy}), the approaches reviewed in this section belong to the \ac{vil} category.

\change{A block scheme that implements \ac{vil} is shown in \Figref{fig:variable_impedance_learning}. Here, the learning algorithm uses $N$ demonstrations in the form $\{\{\bm{x}_{t,n}, \dot{\bm{x}}_{t,n}, \bm{f}_{t,n}^e\}_{t=1}^{T}\}_{n=1}^{N}$ to learn parameterized---with a set of parameters $\boldsymbol{\theta}$---impedance gains. In other words, \ac{vil} approaches learn a (non-linear) mapping $\boldsymbol{\phi}(\cdot)$ in the form\footnote{\change{The mapping for the rotational impedance parameters can be expressed in a similar way.}}
	\Change{
		\begin{align}
		\hat{\mathbf{K}}_t^{\mathcal{P}} &= \boldsymbol{\phi}^{\mathcal{K}}(\bm{x}_t,\dot{\bm{x}}_t,\fet,\boldsymbol{\theta}^{\mathcal{K}}) \label{eq:msd_trans_vil}\\
		\hat{\mathbf{D}}_t^{\mathcal{V}} &=  \boldsymbol{\phi}^{\mathcal{D}}(\bm{x}_t,\dot{\bm{x}}_t,\fet,\boldsymbol{\theta}^{\mathcal{D}})
		\label{eq:msd_ori_vil}
		\end{align}
		where $\hat{\mathbf{K}}_t^{\mathcal{P}}$ and $\hat{\mathbf{D}}_t^{\mathcal{V}}$ represent the \textit{desired} variable stiffness and damping respectively. At run time, the desired variable impedance gains are retrieved from the current measurements (position, velocity, and force) using the learned model in~\eqref{eq:msd_trans_vil}--\eqref{eq:msd_ori_vil}. As shown in~\Figref{fig:variable_impedance_learning}, the desired impedance gains are online processes--the gains are saturated or their rate of change is slowed down--to ensure desired closed-loop properties like stable interactions~\citep{Ficuciello2015variable}.
		Depending on the application, the learned parameters for stiffness $\boldsymbol{\theta}^{\mathcal{K}}$ and damping $\boldsymbol{\theta}^{\mathcal{D}}$ may differ or not.}
	The technique used to approximate the non-linear mappings $\boldsymbol{\phi}^{\mathcal{K}}(\cdot)$ and $\boldsymbol{\phi}^{\mathcal{D}}(\cdot)$ distinguishes the different \ac{vil} approaches.
	
	Training data are typically provided by an expert user, e.g., via kinesthetic teaching as in~\cite{Kronander2014}, and are independent from the underlying controller. At run time, the desired variable impedance gains are retrieved from the learned model and a \ac{vic} (see \secref{sec:vic}) is used to obtain the desired impedance behavior. Note that, in the \ac{vic} block in \Figref{fig:variable_impedance_learning}, there is no connection between the impedance adaptation strategy and the inertia matrix $\bm{M}_t$. This is because most of the approaches for \ac{vil} learn only variably stiffness and damping matrices, as described by \eqref{eq:msd_trans_vil} and \eqref{eq:msd_ori_vil}. On the contrary, several \ac{vic} approaches also perform inertia shaping as indicated in \Figref{fig:variable_impedance_control}. 
}

%\subsection{\texorpdfstring{\ac{il}}{}}

\subsection*{\texorpdfstring{\ac{vil}}{} via Imitation Learning}
%\subsection{}
\label{sec:vil:imitation_learning}

\ac{il} or \ac{lfd} methods are tools to give machines the ability to mimic human behavior to perform a task \citep{hussein2017imitation, ravichandar2020recent}. In this vein, \ac{lfd} is a userfriendly and intuitive methodology for non-roboticists to teach a new task to a robot. In this case, task-relevant information is
extracted from several demonstrations. Standard LfD approaches have focused on trajectory-following tasks, however, recent developments have extended robot learning capabilities to impedance domain \citep{AbuDakka2015, abu2018force, abudakka2020}.

\cite{Kormushev2011} encoded position and force data into a time-driven \ac{gmm} to later retrieve
a set of attractors in Cartesian space through least-squares regression. Stiffness matrices were estimated using the residuals
terms of the regression process. 
\cite{Kronander2014} used kinesthetic demonstrations to teach haptic-based stiffness variations to
a robot. They estimated full stiffness matrices for given positions
using \ac{gmr}, which used a \ac{gmr} that encoded robot Cartesian positions and the Cholesky vector of the stiffness matrix. 
\cite{Saveriano2014learning} follow a similar approach, but exploit the constraints derived by \citep{Khansari2011learning} to guarantee the convergence of the trajectory retrieved via \ac{gmr}.
\cite{Li2014} omitted the damping term from the interaction model and used
\ac{gmm} to encode the pose of the end-effector. Then they found the
impedance parameters and reference trajectory using optimization techniques. \change{\cite{suomalainen2019improving} exploit \ac{lfd} to learn motion and impedance parameters of two manipulators performing a dual-arm assembly. In their evaluation, they show that adapting the impedance of both robots in both rotation and translation is beneficial since it allows to fulfill the assembly task faster and with less joint motions.}

%Unlike previous approaches, our approach learns full stiffness matrices that rely not only on the task dynamics but also on sensed interaction forces.

\cite{Rozo2013} proposed a
framework to learn stiffness in a cooperative assembly task based
on visual and haptic information. They used \ac{tpgmm} to estimate stiffness via \ac{wls} and the Frobenius
norm, where each Gaussian component of the \ac{gmm}
was assigned an independent stiffness matrix. Later, they reformulated their stiffness estimation method as a convex optimization problem, so that optimal
stiffness matrices are guaranteed \citep{Rozo2016}. 

Although traditional \ac{lfd} approaches tend to teach manipulation skills to robots from human expert, \cite{peternel2017robots} proposed a learning method based on \ac{dmp} where a novice robot could learn variable impedance behaviour from an expert robot through online collaborative task execution.

%\subsection{Learning manifolds}
%\label{sec:vil:manifolds}

In \ac{il}, multidimensional data are typically stacked into vectors, de facto neglecting the underlying structure of the data. 
Novel \ac{lfd} approaches explicitly take into account that training data are possibly generated by certain Riemannian manifolds with associated metrics. % on Riemannian metrics show promising results.
Recall that, full stiffness and damping matrices are \ac{spd} (equations \eqref{eq:msd_trans}--\eqref{eq:msd_ori_vil}) so that $\Kpt,\Dvt,\Kot,\Dwt \in \spd$, where $\spd$ is the space of $m \times m$ \ac{spd} matrices. This implies that impedance gains have specific geometric constraints which need special treatment in the learning algorithms. All aforementioned approaches needed to process impedance matrices before and after the learning takes place. Thus, we need to learn directly variable impedance matrices without any reparametrization.

\cite{abu2018force} proposed an \ac{lfd} framework to learn force-based variable stiffness skills. Both forces and stiffness profiles were probabilistically encoded using tensor-based \ac{gmm}/\ac{gmr} \cite{Jaquier2017} without any prior reparametrization. They compared their results with the traditional Euclidean-based \ac{gmm}/\ac{gmr} \cite{calinon2016tutorial} after reparametrizing stiffness matrices using Cholesky decomposition. Their results showed that direct learning of \ac{spd} data using tensor-based \ac{gmm}/\ac{gmr} provides more accurate reproduction than reparametrizing the data and using traditional \ac{gmm}/\ac{gmr}
Two years later, \cite{abudakka2020} reformulated \acp{dmp} based on Riemannian metrics, such that the
resulting formulation can operate with \ac{spd} data in the
\ac{spd} manifold. Their formulation is capable to adapt to
a new goal-\ac{spd}-point.

\section{Variable Impedance Learning Control \texorpdfstring{\change{(VILC)}}{}}
\label{sec:vilc}

\ac{vil} often depends on the underlying control strategy up to the point where defining a clear boundary between the learning algorithm and the controller design becomes impossible. Such approaches belong to the \ac{vilc} category in \Figref{fig:taxonomy} and are reviewed in this section.  

\begin{figure}[t]
	\centering
	\includegraphics[width=0.9\columnwidth]{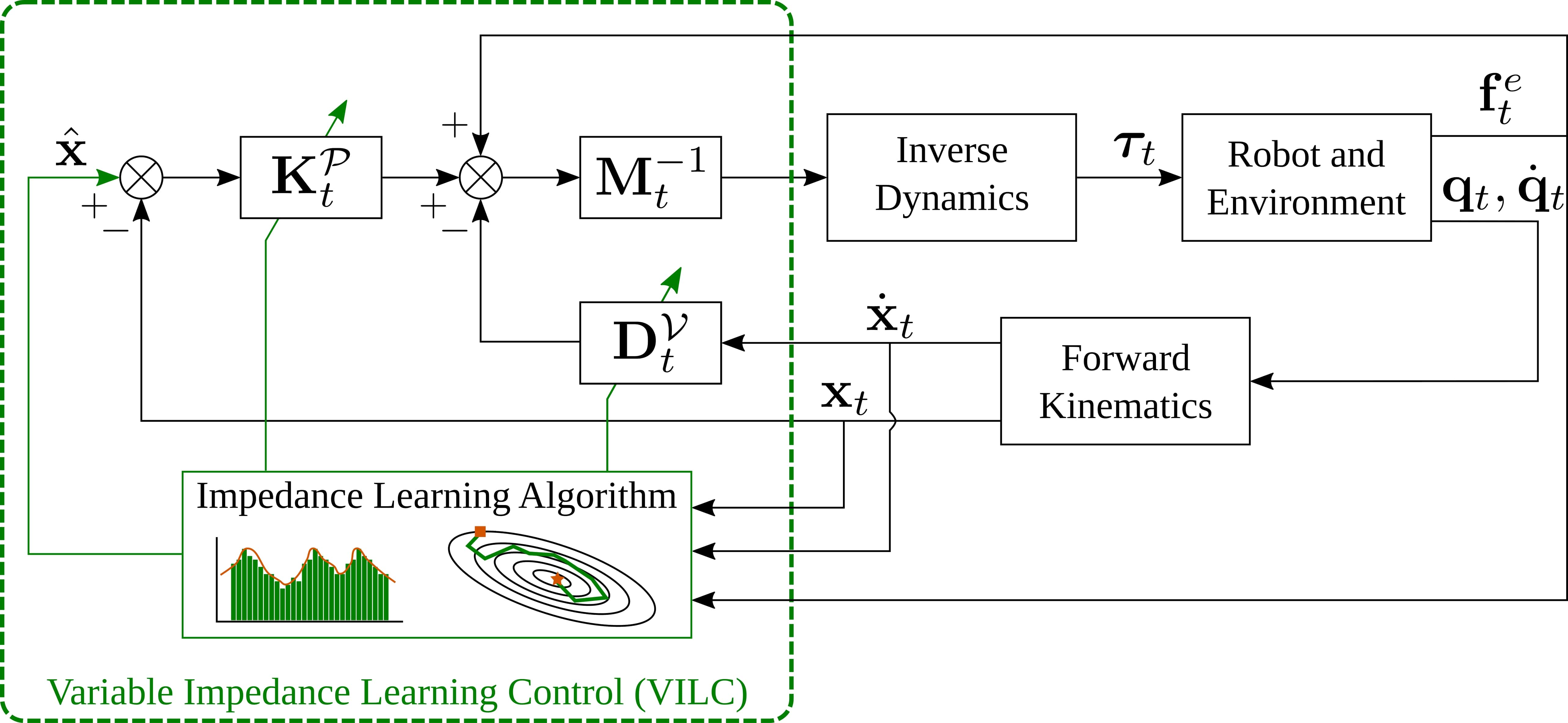}
	\caption{\change{Block scheme of  \ac{vilc}, obtained assuming that $\dot{\hat{\mathbf{x}}} = \ddot{\hat{\mathbf{x}}}  = \mathbf{0}$.}}
	\label{fig:variable_impedance_Learning_control}
\end{figure}

\change{A block scheme that implements \ac{vilc} is shown in \Figref{fig:variable_impedance_Learning_control}. As for \ac{vil}, the learning algorithm uses training data in the form $\{\{\bm{x}_{t,n}, \dot{\bm{x}}_{t,n}, \bm{f}_{t,n}^e\}_{t=1}^{T}\}_{n=1}^{N}$ to learn parameterized---with a set of parameters $\boldsymbol{\theta}$---impedance gains. The key difference between \ac{vil} and \ac{vilc} is that in \ac{vilc} the data collection process itself depends on the underlying control structure. Therefore, the learning and control block are merged in \Figref{fig:variable_impedance_Learning_control}, while they are separated in \Figref{fig:variable_impedance_learning}. Compared to standard \ac{vic}, \ac{vilc} approaches adopt more complex impedance learning strategies requiring iterative updates and/or robot self-exploration. Moreover, \ac{vilc} updates also the target pose (or reference trajectory) while typically rely on constant inertia matrices.}

\subsection{\texorpdfstring{\ac{vilc}}{} via Imitation Learning}
%\subsection{Imitation Learning}
\label{sec:vilc:imitation}

Some of the imitation learning approaches focus on fitting variable impedance gains from training data independently of the way the resulting behavior is executed on real robots. In our taxonomy, shown in \figref{fig:taxonomy}, we have categorized these work as methods for \textit{impedance learning} and reviewed prominent ones in \secref{sec:vil:imitation_learning}. Nevertheless, approaches for imitation learning exist where learning and control tightly integrate and cannot be decoupled. Prominent approaches are reviewed in this section.

\cite{Calinon2010} propose an active learning control strategy to estimate variable stiffness from the inverse of the observed position covariance
encapsulated in a \ac{gmm}. Their approach is limited to tasks displaying variability in position trajectories across demonstrations,
which does not arise in scenarios where the end-effector is constrained to follow a single Cartesian path (e.g., valve-turning tasks).
The Integrated MOtion Generator and Impedance Controller (i-MOGIC) proposed by \citep{Khansari2014modeling} derives the robot trajectory and variable impedance gains from a \ac{gmm} and use them to compute the control input
\begin{change2}
	\begin{equation}
	\bm{u}_t = \sum_{g=1}^G h_g(\bm{x}_t,\dot{\bm{x}}_t)\left[\bm{K}^{\mathcal{P}}_g(\hat{\bm{x}}_g - \bm{x}_t) + \bm{D}^{\mathcal{V}}_g(\dot{\hat{\bm{x}}}_g - \dot{\bm{x}}_t) + \hat{\bm{u}}_g \right].
	\label{eq:i-mogic}
	\end{equation}
	%In \eqref{eq:i-mogic}, 
	where $G$ is the number of Gaussian components, $h_g(\bm{x}_t,\dot{\bm{x}}_t)$ are $G$ state dependent mixing coefficients, $\hat{\bm{x}}_g$ and are $\dot{\hat{\bm{x}}}_g $ local position and velocity targets, $\bm{K}^{\mathcal{P}}_g$ and $\bm{D}^{\mathcal{V}}_g$ are full stiffness and damping matrices, and $\hat{\bm{u}}_g$ are eventual force (spring) preloads. 
\end{change2}
In this formulation, both the trajectory and the impedance gains depend on the robot's state (position and velocity) and are retrieved at run time using sensed information. The stability of the overall closed loop system composed by the robot and the i-MOGIC is proved using Lyapunov arguments.

\change{\cite{Azad2019online} implemented an \ac{il}-based forward model approach with incremental adaptation capability of a state-dependent, time-independent impedance parameters. Moreover, their approach includes a hybrid force-motion controller that provides compliance in particular directions while adapting the impedance in other directions. Recently, \cite{Parent2020Variable} proposed an approach that takes the advantage of the variability that comes from human demonstrations to adapt the stiffness profile according to the precision required in each phase of motion. Their results show a suitable trade-off between precision and compliance.}

%\atM{This approaches use control to let the human show a demonstration online via kinesthetic teaching}
\subsubsection*{Human-in-the-loop}
\label{sec:vilc:hitl}
Kinesthetic teaching is a well-known approach for imitation learning where the human teacher physically guides the robot to demonstrate the task. Kinesthetic teaching is typically applied ``off-line'' to collect motion trajectories by compensating the gravitational torque acting on the robot joints that allows for physical guidance. However, some work extend the kinesthetic teaching framework to provide ``on-line'' demonstrations that are used to adjust the task execution \citep{Lee2011incremental, Saveriano2015incremental, Kastritsi2018progressive, Dimeas2020progressive}. 

In this respect, 
\cite{Lee2011incremental} exploited a variable stiffness profile to generate different impedance behavior in different parts of the state-space. Close to the reference trajectory the stiffness is high to allow accurate tracking. As the difference between reference and executed trajectories increases, for example because an external force is applied to provide a corrective demonstration, the stiffness smoothly decreases to ease the teaching. Finally, if the tracking error exceeds a certain bound, the stiffness grows again. Overall, the approach allows for a local refinement around a nominal trajectory. 

\cite{Saveriano2015incremental} proposed a unified framework for on-line kinesthetic teaching of motion trajectories for both the end-effector and the null-space. External forces arising from kinesthetic teaching are converted into velocity commands with a standard admittance control. This velocity is added to a stack of tasks with variable priority order and executed using the prioritized inverse kinematics approach by \citep{An2015prioritized}. The variable priority order is needed to select the task to execute in case of conflicts, for example when the user tries to distract the end-effector from the nominal trajectory. \cite{Kastritsi2018progressive} used variable stiffness control to allow a human operator to safely interact with the robot during the operation and provide corrective demonstrations, while guaranteeing the overall passivity. They named this incremental learning with human-in-the-loop \textit{progressive automation} since the robot ``progressively'' becomes autonomous as the number of iterations grows and the tracking error decreases. The framework has been further extended to adapt periodic movements via human--robot interaction \citep{Dimeas2020progressive}.

An impedance behavior, defined as in \eqref{eq:msd_trans_vic}--\eqref{eq:msd_ori_vic}, has an intrinsic \textit{multi-modal nature} since it consists of a reference trajectory and (variable) impedance gains. \cite{Peternel2014, Peternel2018, Yang2018dmp, Wu2019variable} designed multi-modal interfaces to let the human to explicitly teach an impedance behavior to the robot. More in details, \cite{Peternel2014} used a marker attached to the hand to measure the position and \ac{emg} sensors to capture the muscular activation of the arm, where high values of the \ac{emg} signal are mapped into high robot stiffness. Information captured via this multi-modal interface are used in an on-line learning loop and used to update the parameters of two \acp{dmp}, one used to generate the desired trajectory and one the desired stiffness. 

Similarly, \cite{Yang2018dmp} used \ac{emg} and a master robotic arm to demonstrate variable impedance behaviors to a slave robotic arm. \cite{Wu2019variable}, instead, combined \ac{emg} measurements and the forces sensed during kinesthetic teaching into an admittance controller with variable impedance gains.
A limitation of these work is that they require a complex setup and a long calibration procedure to achieve good performance. \cite{Peternel2018} designed a multi-modal interface consisting of a haptic device that measures the arm trajectory and return a force feedback and a potentiometer that the user press to modulate the stiffness. As in their previous work \citep{Peternel2014}, sensed information are used to on-line estimate the \acp{dmp} parameters. The calibration procedure to make the multi-modal interface \citep{Peternel2014} has been simplified later for easier use \citep{peternel2016towards,peternel2017human}.
%\cite{peternel2016towards,peternel2017human} simplify the calibration procedure to make the multi-modal interface of \citep{Peternel2014} easier to use. 
Finally, \cite{peternel2018robot} further extended their control approach for multi-modal interface \citep{peternel2016towards,peternel2017human} by using position and force feedback as well as muscular activity  measurements. The latter is used to estimate the human physical fatigue and then teach the robot (based on \ac{dmp}) to reduce human's fatigue by increasing its contribution to the execution of the cooperative task. A a result, the robot gradually takes over more effort when the human gets tired.

\subsection{\texorpdfstring{\ac{vilc}}{} via Iterative Learning}
Tuning variable impedance gains can be seen as an repeated learning process where the robot improves its performance at each iteration. The idea of a repeated learning results in two category of approaches, namely those based on \textit{iterative learning} and those based on  \textit{\ac{rl}}. Iterative learning approaches are reviewed in this section, while \ac{rl} is covered in \secref{sec:vilc:control_RL}. 

A bunch of work \citep{Cheah1998learning, Gams2014coupling, Uemura2014iterative, AbuDakka2015, Kramberger2018passivity} that propose to iteratively adjust the impedance rely on the \ac{ilc} framework~\citep{Bristow2006survey}. \ac{ilc} assumes that the performance of an agent that repeatedly performs the same task can be improved by learning from past executions. In the conventional \ac{ilc} formulation, the objective is to reduce the trajectory tacking error while rejecting periodic disturbances. This is obtained by adjusting the pre-defined control input with a corrective term that linearly depends on the tracking error, i.e.
\begin{change2}
	\begin{equation}
	\bm{u}_{r+1,t} = \bm{u}_{r,t} + \gamma_r \bm{e}_{r,t}.
	\label{eq:ilc_standard}
	\end{equation}
\end{change2}
\change{where the subscript $r$ indicates the iteration number, while $t$ a time dependency, $\bm{u}_{r,t}$ is the control input to adjust, and $\bm{e}_{r,t}$ the trajectory tracking error. The gain $\gamma_r$ in~\eqref{eq:ilc_standard} is} iteratively updated in a way that ensures asymptotic convergence of the tracking error at least if the system to control has a linear dynamic. \Change{The conventional \ac{ilc} described by~\eqref{eq:ilc_standard} relies on the trajectory tracking error and it is not directly applicable to \ac{vilc}. The easiest way to use \ac{ilc} to reproduce impedance behaviors is to combine it with a properly designed \ac{vic}. However, this approach does not allow the robot to learn suitable impedance gains. In order to learn variable impedance gains, the error term in ~\eqref{eq:ilc_standard} needs to be modified to describe the discrepancy between the desired and the real impedance behavior. Common strategies exploited in \ac{vilc}} to modify the conventional \ac{ilc} formulation are described as follows.

Recall that the goal of impedance control is to let the robot behave as the second order dynamics specified in \eqref{eq:msd_trans_vic}--\eqref{eq:msd_ori_vic}. By specifying the desired trajectory and impedance gains, the desired dynamics in \eqref{eq:msd_trans_vic}--\eqref{eq:msd_ori_vic} becomes a target impedance behavior and \ac{ilc} \eqref{eq:ilc_standard} can be used to enforce the convergence of the robot behavior to this target. \cite{Cheah1998learning} combined a standard, PD-like controller with a learned feedforward term and a dynamic compensation term. The feedforward term is update with an iterative rule and a proper selection of the compensator gains ensures the convergence of the iterative control scheme to the target behavior. The \ac{ilc} scheme in \citep{Cheah1998learning} relies on a standard impedance controller and requires  measurement of interaction forces and a fixed target impedance behavior, i.e. constant gains and a predefined desired trajectory. As a consequence, the interaction may become unstable if the environment changes significantly. This undesirable effect is overcome by the biomimetic controller \citep{Yang2011human} and inspired by experimental observation that humans optimize the arm stiffness to stabilize point-to-point motions \cite{Burdet2001}. In this case, the iterative update rule, not derived from \ac{ilc} but from human observation, involves both feedforward force and impedance gains. Notably, no force measure is required to implement this iterative control scheme that guarantee stable interactions and error cancellation with a well-defined margin.

There is an intrinsic advantage in adapting the reference trajectory during the motion that is not exploited by \citep{Cheah1998learning,Yang2011human}. Indeed, by modulating the desired trajectory, one can let the robot anticipate contacts before they occur. This potentially leads to stable transitions between interaction and free motion phases and allows for adaptation to changing environments. This possibility is investigated by \citep{Gams2014coupling}. \citeauthor{Gams2014coupling} proposed to couple two \acp{dmp} with an extra force term. The approach is relatively general since the coupling term may represent different reference behaviors including desired contact forces or relative distance between the robot hands. The forcing term is updated from sensory data using an \ac{ilc} scheme with guaranteed convergence. 
Varying the joint trajectory as well as the gains of a diagonal stiffness matrix with an \ac{ilc} rule is exploited by \citep{Uemura2014iterative} to generate energy-efficient motions for multi-joint robots with adjustable elastic elements. The approach has guaranteed convergence properties and can effectively handle boundary conditions like desired initial and goal joint position and velocity. 

\cite{AbuDakka2015} iteratively modified the desired positions and orientations to match forces and torques acquired from human demonstrations. While \citep{Gams2014coupling} considered only positions, \citep{AbuDakka2015} combined unit quaternions, a singularity-free orientation representation, and \ac{dmp} to represent the full Cartesian trajectory. Moreover, their \ac{ilc} scheme learns how to slow down the \ac{dmp} execution such that the force/torque error is minimized. The approach is experimentally validated on challenging manipulation tasks like assembly with low relative tolerances.

As discussed in \secref{sec:vilc:imitation} and \citep{Van2000l2}, passivity is a powerful tool to analyze the stability of the interaction with a changing and potentially unknown environment. 
\citep{Kramberger2018passivity} propose an admittance-based coupling of \ac{dmp} that allows both trajectory and force tracking in changing environments. The paper introduces the concept of \textit{reference power trajectory} to describe the target behavior of the system under control---consisting of \ac{dmp}, robot, and passive environment. Using a power observer, the reference power error is computed and used in an \ac{ilc} scheme to learn a varying goal of the \ac{dmp}. As a result, the varying goal reduces the trajectory and force tracking errors while maintaining the passivity of the overall system.

\subsection{\texorpdfstring{\ac{vilc}}{} via Reinforcement Learning}
\label{sec:vilc:control_RL}
%\atF{\citep{Kim2010, Buchli2011,Dimeas2015reinforcement,Rey2018,Martin2019variable} exploit the standard impedance/admittance control law and learn reference trajectories and variable gains gains. Is it the case to consider them as learning approaches? Maybe a bit more discussion on what we mean with learning control is needed.}
\ac{rl} is a widely studied topic in the learning and control communities, and it is beyond the scope of this survey to provide an exhaustive description of the topic. For the interested reader, \citep{Sutton2018reinforcement} is a good reference to start with \ac{rl}. \cite{Kormushev2013reinforcement,Kober2013reinforcement,Deisenroth2013survey} described robotic specific problems of \ac{rl}. \citep{Chatzilygeroudis20survey} reviews recent advancement in the field with a particular focus on data-efficient algorithms, while \citep{Arulkumaran2017deep} focuses on deep learning solutions. Instead, we assume that the reader is already familiar with \ac{rl} and focus on presenting existing approaches for \ac{rl} of variable impedance control. 

In interaction tasks, variable impedance (or admittance) control can be adopted as a parameterized policy in the form\footnote{\change{The parameterization~\eqref{eq:vilc_rl} can be applied in joint or Cartesian spaces and extended to consider the orientation.}}
\begin{change2}
	\begin{equation}
	\bm{\pi}_{\theta,t} = \bm{K}^{\mathcal{P}}_{\theta,t}(\hat{\bm{x}}_{\theta,t} - \bm{x}_t) + \bm{D}^{\mathcal{V}}_{\theta,t}(\dot{\hat{\bm{x}}}_{\theta,t} - \dot{\bm{x}}_t) + {\bm{f}}_t^e,
	\label{eq:vilc_rl}
	\end{equation}
	where $\bm{\pi}_{\theta,t}$ is a control policy depending on a set of learnable paramters $\bm{\theta}$. The parameters $\bm{\theta}$ define the desired trajectory ($\hat{\bm{x}}_{\theta,t}$ and $\dot{\hat{\bm{x}}}_{\theta,t}$) as well as the desired impedance (or admittance) behavior ($\bm{K}^{\mathcal{P}}_{\theta,t}$ and $\bm{D}^{\mathcal{V}}_{\theta,t}$). These parameters can be
\end{change2}
optimally tuned using approaches from \ac{rl} \citep{Kim2010, Buchli2011, Dimeas2015reinforcement, Rey2018, Martin2019variable}. Experiments show that adopting such a specialized policy results in increased sample efficiency and overall performance  in complex interaction  tasks like contact-rich manipulation.

More in details, \cite{Kim2010} used an episodic version of the Natural Actor-Critic algorithm \citep{Peters2008natural} to learn a variable stiffness matrix. Their algorithm targets planar $2$-link manipulators since the $2\times 2$ \ac{spd} stiffness matrix is completely represented by $3$ scalar values, namely the magnitude, the shape, and the orientation. This keeps the parameter space small and allows for a quick convergence to the optimal stiffness. However, the effectiveness of the approach in realistic cases, e.g. a spatial manipulator with $6$ or $7$ links, is not demonstrated. 

\cite{Buchli2011} used the \ac{pi2} algorithm \citep{Theodorou2010generalized} to search for the optimal policy parameters. A key assumption of \ac{pi2} is that the policy representation is linear with respect to the learning parameters. Therefore, \cite{Buchli2011} proposed to represent the desired position and velocity as a \ac{dmp} \citep{Ijspeert2013dynamical}, \begin{change2}
	a policy parameterization that is linear with respect to the learning parameters. For the stiffness, authors exploited a diagonal stiffness matrix and express the variation (time derivative) of each diagonal entry as 
	\begin{equation}
	\dot{k}_{\theta_j,t} = \alpha_j \left(\bm{g}_j\trsp(\bm{\theta}_j+\bm{\epsilon}_{j,t})- k_{\theta_j,t} \right), \quad j=1,\ldots,J ,
	\label{eq:pi2_stiffness}
	\end{equation}
	where $j$ indicates the $j$-th joint, $k_{\theta_j,t}$ is the stiffness of joint $j$, $\bm{\epsilon}_{j,t}$ is a time-dependant exploration noise, $\bm{g}_j$ is a sum of $G$ Gaussian basis functions, and $\bm{\theta}_j$ are the learnable parameters for joint $j$. The stiffness parameterization in~\eqref{eq:pi2_stiffness} is also linear in the parameters and \ac{pi2} can be applied to find the optimal policy. 
\end{change2}
It is worth noticing that \cite{Buchli2011} used a diagonal stiffness matrix and one \ac{dmp} for each motion direction (or joint), allowing the \ac{pi2} to optimize the behavior in each direction independently. This has the advantage of reducing the parameter space and, possibly, the training time. However, a diagonal stiffness neglects the mutual dependencies between different motion directions which may be important depending on the task. Following a similar idea of~\cite{Buchli2011},~\cite{Rey2018} parameterized the policy as a non-linear, time invariant dynamical system using the \ac{seds}~\citep{Khansari2011learning}. This is a key difference with the work by~\cite{Buchli2011}, since \ac{dmp} introduces an explicit time dependency. The idea of \ac{seds} is to encode a first order dynamics into a \ac{gmm} 
\begin{change2}
	\begin{equation}
	\dot{\bm{\xi}}_t = \sum_{g=1}^G h_g(\bm{\xi}_t)\bm{A}^{\mathcal{P}}_g(\hat{\bm{\xi}} - \bm{\xi}_t),
	\label{eq:seds}
	\end{equation}
	where $\bm{\xi}_t$ is a generic state variable, $\hat{\bm{\xi}}$ the goal state, $0 < h_g(\bm{x}_t) \leq 1$ are $G$ state-dependent mixing coefficients and the matrices $\bm{A}^{\mathcal{P}}_g$ depend on the learned covariance. Using Lyapunov theory~\citep{Slotine1991applied}, authors conclude that the system~\eqref{eq:seds} globally converges to $\hat{\bm{\xi}}$ if all the $\bm{A}^{\mathcal{P}}_g$ are positive definite\footnote{\change{In the original formulation the matrices $\bm{A}^{\mathcal{P}}_g$ are assumed negative definite~\citep{Khansari2011learning}. We have slightly modified the system~\eqref{eq:seds} to use positive definite matrices and be consistent with other equations in this survey.}}.    
\end{change2}
The \ac{pi2} algorithm is modified accordingly to fulfill the \ac{seds} stability requirements during the exploration.  Authors also propose to \change{augment the state vector $\bm{\xi}$ to include position and stiffness, and to encode a variable stiffness profile using~\eqref{eq:seds}}. A variable impedance controller is then used to perform interaction tasks where the variable, diagonal stiffness matrix and the reference trajectory are retrieved at each time step from the learned dynamical system.

\cite{Dimeas2015reinforcement} adopted an admittance control scheme with a constant inertia matrix and null stiffness, and exploits Fuzzy Q-learning \citep{Berenji1994fuzzy, Jouffe1998fuzzy} to discover optimal variable damping gains (one for each motion direction). The goal of the learning agent is to minimize the robot jerk (third time derivative of the position) in human-robot co-manipulation tasks. Authors conduct a user study with $7$ subjects performing co-manipulation task with a real robot, showing that their approach converges in about $30$ episodes to a sub-optimal policy that reduces both time and the energy required to complete the task.

As already mentioned, approaches in \citep{Buchli2011, Rey2018} rely on a diagonal stiffness matrix to reduce the parameter space and the corresponding search time. The drawback is that a diagonal stiffness neglects the inter-dependencies between different motion directions. This problem is faced by \citep{Kormushev2010robot}. \citeauthor{Kormushev2010robot} proposed to learn an acceleration (force) command as a mixture of $G$ proportional-derivative systems
\begin{change2}
	\begin{equation}
	\bm{u}_t = \sum_{g=1}^G h_{g,t}\left[\bm{K}_g^{\mathcal{P}}(\hat{\bm{x}}_g - \bm{x}_t)-d^{\mathcal{V}}\dot{\bm{x}}_t\right], 
	\label{eq:korm_accel}
	\end{equation}
	where $\hat{\bm{x}}_g$ are $G$ local target, $h_{g,t}$ are time varying mixing coefficients, $d^{\mathcal{V}}$ is a constant damping, and $\bm{K}_g^{\mathcal{P}}$ \end{change2} are full stiffness matrices (that authors call \textit{coordination matrices} since they describe the local dependency across different motion directions). Clearly, the control input $\bm{u}_t$ in \eqref{eq:korm_accel} realizes a variable impedance control law. The method is applied to the highly-dynamic task of flipping a pancake. 
The work by \citep{Luo2019reinforcement} follows a different strategy to search for a force control policy. 
The iterative Linear-Quadratic-Gaussian approach \citep{Todorov2005generalized} is used to find a time-varying linear-Gaussian controller representing the end-effector force. \change{In this case, injecting a Gaussian noise in the control input is beneficial since it helps to reduce the model bias of the RL algorithm \citep{Deisenroth2015gaussian}.} The generated force is inserted into an hybrid position/force controller that implicitly realize an adaptive impedance behavior, i.e. the robot has high (low) impedance in free (contact) motion. A neural network is trained in a supervised manner to represent and generalize the linear-Gaussian controller, as experimentally demonstrated in $4$ assembly tasks.

In principle, most of (probably all) the approaches developed for robot control can be used to map a learned policy into robot commands. \cite{Martin2019variable} presented an interesting comparison between well-known controllers used to map a policy into robot commands. Clearly, the output of the policy depends on the chosen controller. For example, the policy of a joint torque controller outputs the desired torque. In case of a Cartesian variable impedance controller the policy output are the desired pose, velocity, damping and stiffness. They compared $5$ popular controllers, namely joint position, velocity, and torque, and Cartesian pose and variable impedance, on $3$ tasks (path following, door opening, and surface wiping). The comparison considers the following metrics: \textit{i)} sample efficiency and task completion, \textit{ii)} energy efficiency, \textit{iii)} physical effort (wrenches applied by the robot), \textit{iv)} transferability to different robots, \textit{v)} sim-to-real mapping. Their findings show that the Cartesian variable impedance control performs well for all the metrics. Interestingly, a variable impedance control policy is easier to port to another robot and can be almost seamlessly transfer from a simulator to a real robot.

\ac{rl} methods have great potential and are effective in discovering sophisticated control policy. However, especially for \textit{model-free} approaches, the policy search can potentially be extremely data inefficient. One possibility is to alleviate this issue is to use a ``good'' initial policy and locally refine it. Imitate the human impedance behavior is a possibility to initialize the control policy and standard \ac{rl} techniques or, more effectively, inverse \ac{rl} approaches can be used to refine the initial policy \citep{Howard2013transferring}.  
Alternatively, there exists a class of \textit{model-based} \ac{rl} approaches that is intrinsically data-efficient \citep{Sutton2018reinforcement}. Loosely speaking a model-free learner uses an approximate dynamic model, learned from collected data, to speed up the policy search.  

In the context of \ac{vic}, \cite{Li2018efficient, li2019efficient} used Gaussian processes \citep{Williams2006gaussian} to learn a probabilistic representation of the interaction dynamics. \change{In order to overcome the measurement noise of the force/torque sensor, \cite{Li2018efficient} designed a Kalman filter to estimate the actual interaction forces.} The learned model is used to make long term reward prediction and optimize the policy using gradient-based optimization as originally proposed by \citep{Deisenroth2015gaussian}.
Gaussian process are extremely sample-efficient. However, they do not scale with large datasets and tend to smooth out discontinuities that are typical in interaction tasks. In order to realize sample and computationally efficiency, \cite{Roveda2020model} proposed a mode-based \ac{rl} framework that combines \ac{vic}, an ensemble of neural networks to model human--robot interaction dynamics, and an online optimizer of the impedance gains. The ensemble of networks, trained off-line and periodically updated, is exploited to generate a distribution over the predicted interaction that reduces the overfitting and captures uncertainties in the model.

\subsubsection*{Stability in \texorpdfstring{\ac{vic}}{} exploration}\label{subsubsec:vilc_stability}
Realizing a safe exploration that avoids undesirable effects during the learning process is a key problem in modern \ac{ai} \citep{Amodei2016concrete}. For \ac{rl}, this is particularly important in the first stages of the learning when the agent has limited knowledge of the environment dynamics and applies control policies that are potentially far from the optimal one. The aforementioned approaches are promising and they can potentially discover complex variable impedance policies. However, none of them is designed to guarantee a safe exploration. A possible way to guarantee a safe exploration is to identify a set of safe states where the robot is stable (in the sense of Lyapunov) and to prevent the robot to visit unsafe states \citep{Berkenkamp2017safe, Chow2018lyapunov, Cheng2019end}. With the goal of guaranteeing Lyapunov stability during the exploration, \cite{Khader2020stability} proposed an all-the-time-stability exploration strategy that exploits the i-MOGIC policy parameterization in~\eqref{eq:i-mogic}. As detailed in \secref{sec:vilc:imitation}, i-MOGIC allows to learn a \ac{vic} with guaranteed Lyapunov stability.  As a difference with the \ac{seds}-based approach by \citep{Rey2018}, the i-MOGIC parameterization allows to learn full stiffness and damping matrices that encode the synergies between different motion directions. The stability constraints derived by \citep{Khansari2014modeling} are exploited by \citep{Khader2020stability} to constraint the parameters during the policy updates, guaranteeing a stable exploration.  

The idea of using stable dynamical systems to parameterize a control policy is promising, since it allows for a stable exploration. However, it is not clear if an optimal policy can be found in the constrained parameter spaces. At this point, further investigation is required to quantify the limitations introduced by the specific policy parameterizations. A possible solution could be to simultaneously update the policy parameters and the Lyapunov function. This would allow to relax the stability constraints by increasing both the safe set and, as a consequence,  the probability of finding an optimal policy.

\begin{change2}
	\section{Discussion}
	\label{sec:discussion}
	
	\begin{table}[h]
		\resizebox{\textwidth}{!}{%
			{\renewcommand\arraystretch{1.2} 
				 \centering
				\begin{tabular}{|m{0.02\textwidth}|m{0.1\textwidth}|m{0.43\textwidth}|m{0.43\textwidth}|}
					\noalign{\hrule height 1.5pt}
					\multicolumn{2}{|c}{\textbf{Approach}} & \multicolumn{1}{|c}{\textbf{Advantages}} & \multicolumn{1}{|c|}{\textbf{Disadvantages}} \\
					\noalign{\hrule height 1pt}
					\multirow{5}[54]{*}{\rotatebox[origin=c]{90}{\textbf{Existing}}} &
					%%%%%%STABILITY
					\textbf{Stability and Passivity of VIC} & Model-based solutions, where the models are often simplified computational representations, are efficient and accurate. Guarantee the stability (or passivity) is of importance for safe interactions. & Rely on accurate models of the system under control to  work well. Derive accurate models is, in some cases, nontrivial which complicates the overall design and makes the solution less general. \\ \cline{2-4}
					&  
					%%%%%%LOOP
					\textbf{Human-in-the-Loop} & \change{During the execution, human can react where the AI algorithms are not confident about the next reaction.} Human impedance, estimated via \ac{emg} sensors, postural markers, and/or haptic devices, often represents a good target impedance for the robotic arm. & Require prior knowledge on the human anatomy, a complex setup with multiple sensors, and a long calibration time. \change{Moreover, the system can be influenced by possible human error, in addition to lack of repeatability.} \\ \cline{2-4}
					&  %%%%%%IL
					\textbf{Imitation Learning} %\cite{Calinon2010, Khansari2014modeling, Kastritsi2018progressive, Dimeas2020progressive}  
					& \change{User-friendly and easy learning framework to teach robots.} Humans use variable impedance strategies in many of their daily activities and can naturally demonstrate a proper impedance behavior to solve a specific task.  & \change{The quality of learning can be influenced by the teacher performance. Some tasks are complex enough to be demonstrated.} Directly transfer the impedance policy from a human to a robot is not always possible and may require sophisticated strategies or hand tuning.  \\ \cline{2-4}
					& 
					%%%%%%ILC
					\textbf{Iterative Learning} %\cite{Cheah1998learning, Yang2011human, Gams2014coupling, Uemura2014iterative, AbuDakka2015, Kramberger2018passivity}  
					& These approaches are computationally and data efficient. Convergence to the optimal parameters can be analytically proved. & The target impedance behavior has to be manually defined which makes hard to generalize the approach to dissimilar tasks. \change{Moreover, standard \ac{ilc} assumes  that the system is already stable or stabilized with a suitable controller. Moreover, it needs multiple task repetitions with the same duration and initial conditions, which is hard to guarantee in real scenarios.} 
					\\ \cline{2-4}
					& 
					%%%%%%RL
					\textbf{\ac{rl}} %\citep{Kormushev2010robot, Kim2010, Buchli2011, Dimeas2015reinforcement, Rey2018, Li2018efficient, Martin2019variable, Roveda2020model, Khader2020stability}  
					& The robot may potentially discover control policies to solve complex, hard to model tasks. The usage of a specialized policy paramaterization  increases the data efficiency and the policy transferability.  & Specialized policies, like the ones based on VIC, as well as safety requirements limit the exploration capability of the learning agent increasing the risk to get stuck into a policy far from the optimal one.\\
					\hline \hline
					\rotatebox[origin=c]{90}{\textbf{Envisioned}}  &  \multicolumn{3}{m{\textwidth}|}{
						\begin{itemize}
							\item The ideal impedance behavior should be:
							\begin{itemize}
								\item[--] stable, accurate, and robust like a control approach, without requiring an accurate model or domain specific knowledge like in reinforcement learning.
								\item[--] computational and data efficient, as well as easy to setup.
							\end{itemize}
							\item Enhanced generalization capabilities are also required to adapt the robot behavior to different situations. None of the reviewed approaches has all these features. However, some approaches have great potential and deserve to be further investigation.
							\item Manifold learning has shown interesting performance in learning variable impedance behaviors \citep{abu2018force}. In many applications, not only in impedance learning, the training data below to a certain manifold, but the underlying structure of the data is typically not properly exploited by the learning algorithm. Manifold learning remains a widely unexplored and rather promising topic.
							\item Stability guarantees are a need when the robot interacts with the environment and the safe reinforcement learning formalism seems the route to learn effective impedance policies. The most powerful reinforcement learning are extremely data-greedy that poses several limitations on their applicability. In this context, model-based approaches with stability guarantee seem better suited but their effectiveness has not be fully investigated.  
						\end{itemize}
					}  \\
					\noalign{\hrule height 1.5pt}
				\end{tabular}
		}}
		\caption{A comparison of the main approaches for variable impedance learning and control.} 
		\label{tab:comparison}
	\end{table}
	
	In this paper, we presented a review for the main learning and control approaches used in variable impedance controllers. 
	\tabref{tab:comparison} summarizes the general advantages and disadvantages of these approaches.
	
	As stated in \tabref{tab:comparison}, we envision a framework that inherits features from all the different approaches.
	The ideal framework is to be accurate and robust like a properly designed controller, and, at the same time, flexible and easy to generalize like a learning approach. However, there are several theoretical and practical difficulties that need to be overcome to realize the envisioned framework.

	Theoretical guarantees like stability and robustness become difficult to prove in complex systems like a robot manipulator physically interacting with an unstructured environment. Existing approaches make several simplification assumptions, e.g., interactions with a passive environment, to derive theoretical guarantees. These assumptions significantly restrict the domain of application of developed \ac{vic} approaches. In this respect, passivity theory arises as a promising approach given the relatively general working assumptions (see \secref{sec:vic:stability}). 
	However, the passivity framework, as most control approaches, is model-based and sometimes it is hard to come up with a suitable analytical solution without simplification assumptions. It is evident that control alone cannot solve the problem.  
	
	Learning-based approaches are designed to work in complex scenarios under minimal assumptions. For instance, many model-free \ac{rl} approaches only require a reward on the robot's performance to discover a sophisticated and generalizable control policy. This comes at the cost of long training time and possible unsafe robot behaviors. In general, training time and safety are not always an issue for the learning community, but they represent a clear limitation in robotics. Described work on safe and model-based \ac{rl} (see \secref{sec:vilc:control_RL}) started to address these issues, but results are still preliminary.
	
	It is evident, from the previous discussion on the limitations of learning and control approaches, that \ac{vilc} is the route to realize an omni-comprehensive variable impedance framework. However, this poses further challenges to overcome:
	\begin{itemize}
		\item \ac{il} is a paradigm for teaching robots how to perform new tasks even by a non-programmer teachers/users. In this context, \ac{il} approaches extract task-task relevant information (constraints and requirements) from single/several demonstration(s) which can enable adaptive behavior. The approaches presented in Section~\ref{sec:vil:imitation_learning} show successful examples of how diverse impedance tasks---\eg peg-in-the-hole, assembly, \etc---can be learned via human imitation. However, the simple imitation of the task demonstration(s) is prone to failures, especially when physical interaction is required. Possible reasons to fail include\footnote{\change{An exhaustive discussion about pros and cons of \ac{il} is beyond the scope of this survey. Therefore, we focus on limitations of \ac{il} that particularly affect impedance learning. The interested reader is referred to \citep{ravichandar2020recent} for a general discussion about strengths and limitations of \ac{il}.}}: \textit{i)} poor demonstrations provided by inexpert teachers, \textit{ii)} inaccurate mapping between human and robot dynamics, and \textit{iii)} insufficient demonstrations to generalize a learned task.
		
		To overcome these limitations, one needs to endow robots with the ability to generalize to unseen situations of the task. This generalization can be done by combining demonstration driven approaches like \ac{il} with trial-and-error, reward driven learning (\eg \ac{rl}). 
		\item Policy parameterization is needed to cope with the continuous nature of state and action space in robotics. Moreover, a proper policy representation, like the i-MOGIC used by \cite{Rey2018,Khader2020stability}, may lead to desired properties like the all-the-time-stability described in \secref{subsubsec:vilc_stability}, but further investigations are needed to understand if and how a specific policy parameterization limits the learning capabilities of the adopted algorithm.
		\item Safety of a system is typically ensured by constraining the state-space to a safe (sub-)set. When applied to \ac{rl}, this limits the robot exploration to a certain safe set which maybe be too conservative to discover interesting policies. Moreover, the safe set is typically hand designed by an expert. A possibility to address this issue is to use a very basic safe set (e.g., a ball around the initial state of the robot), and improve the estimation of the safe set during the learning. Recently \cite{wabersich2018linear} have proposed a data-driven approach to iteratively increase the safe set. The approach work only for linear systems and the extension to non-linear ones is, as usual, non-trivial. 
		\item We seek for policies that generalize well and are applicable in a wide range of situations. The generalization capabilities of a learning algorithm often depends on the adopted feature representation. Most of the approaches either use diagonal stiffness and damping matrices or simply vectorize the full matrices to form a training set. However, as discussed in \secref{sec:vil:imitation_learning}, impedance parameters are \ac{spd} matrices and the vectorization simply discards this information. Therefore, a Riemannian manifold represents the natural space from which training data are sampled, and taking the underlying manifold structure often lead to better extracted features that increase the discriminative power and the generalization capabilities of the learning algorithm.
		Recent work \cite{abu2018force, abudakka2020} reformulated the learning problem by taking into account the underlying Riemannian manifold structure and show improved performance compared to standard approaches based on vectorization. Results are promising but too preliminary to definitely assess the generalization capabilities of manifold-based approaches.
	\end{itemize}
	
	\change{Building a safe \ac{rl} algorithm on top of a manifold representation, like \ac{spd}, is, at least in theory, possible. However, at the best of our knowledge, this is still an ongoing research topic and there is no available approach.}
\end{change2}

\section{Concluding Remarks}
\label{sec:conclusion}

%In this survey, we analyzed the state-of-the-art of \ac{vic} approaches 

Varying the robot impedance during the task execution is a popular and effective strategy to cope with the unknown nature of everyday environments. In this survey, we have analyzed several approaches to adjust impedance parameters considering the task at end. Traditionally, variable impedance behavior were achieved by means of control approaches, namely the variable impedance control. More recently, the learning community has also focused on the problem attempting to learn impedance gains from training data (\ac{vil}) or a non-linear controller with varying impedance (\ac{vilc}).
Each approach has its own advantages and disadvantages, that we have summarized in \tabref{tab:comparison}.

At the current stage, none of the approaches has all the features that a variably impedance behavior requires. Control approaches have solid mathematical foundations that make them robust and efficient, but require a significant amount of prior knowledge. Learning approaches may require less amount of prior information, but they are often data and computationally inefficient. \change{These limitations, as discussed in \secref{sec:discussion}, reduce the applicability of variable impedance approaches and have heavily burden the spread of robotic solutions in dynamic and unstructured environments.} 

We believe that manifold and reinforcement learning are the most promising approaches to \change{overcome existing limitations of \ac{vilc} approaches and have the potential to} learn variable impedance behaviors that are effective both in industrial and service scenarios.

\section*{Conflict of Interest Statement}

The authors declare that the research was conducted in the absence of any commercial or financial relationships that could be construed as a potential conflict of interest.

\section*{Author Contributions}
FAD was responsible to review the approaches for impedance control, variable impedance control, and impedance learning. He also gave a major contribution in writing the introduction and the final remarks. 

\noindent MS was responsible for the taxonomy used to categorize approaches for variable impedance learning and control. He was also reviewing the approaches used for variable impedance learning control.

\section*{Funding}
This work is partially supported by CHIST-ERA project IPALM (Academy of Finland decision 326304) and partially supported from the Austrian Research Foundation (Euregio IPN 86-N30, OLIVER).

\bibliographystyle{frontiersinSCNS_ENG_HUMS} 
\bibliography{ref}

\end{document}